\documentclass[letterpaper]{article}
\usepackage{times}
\usepackage{helvet}
\usepackage{amsmath,bm}
\usepackage{courier}
\usepackage{graphicx}
\usepackage{multirow}
\usepackage{CJK}
\usepackage{color,xcolor,colortbl}
\definecolor{mygray}{gray}{.7}
\usepackage{rotating}
\usepackage{array}
\usepackage{cite}
\usepackage{geometry}
\date{}

\title{Two are Better than One:\\
An Ensemble of Retrieval- and Generation-Based Dialog Systems}
\author{\em Yiping Song,$^{1}$ Rui Yan,$^{2}$ Xiang Li,$^{1}$ Dongyan Zhao,$^{2}$ Ming Zhang$^{1}$ \\
${}^{1}$ School of EECS, Peking University, China \\
${}^{2}$ Institute of Computer Science and Technology, Peking University, China \\
\{songyiping, ruiyan, lixiang.eecs, zhaody, mzhang\_cs \}@pku.edu.cn
}
\begin{document}
\begin{CJK*}{UTF8}{gkai}

\newcommand{\circone}{\scalebox{.9}{\Large\textcircled{\small{1}}}}
\newcommand{\circtwo}{\scalebox{.9}{\Large\textcircled{\small{2}}}}

\newcommand{\seqseq}{\texttt{seq2seq}}
\newcommand{\multiseqseq}{\texttt{biseq2seq}}
\newcommand{\rstar}{r^*}
\newcommand{\rplus}{r^\text{+}}

\newcommand\Tsmall{\rule{0pt}{2ex}}
\newcommand\Tstrut{\rule{0pt}{2.4ex}}         
\newcommand\Bstrut{\rule[-0.9ex]{0pt}{0pt}}   

\newcommand{\PreserveBackslash}[1]{\let\temp=\\#1\let\\=\temp}
\newcolumntype{C}[1]{>{\PreserveBackslash\centering}p{#1}}

\def\shortcite{\def\citeauthoryear##1##2{##2}\@icite}
\def\citeauthor{\def\citeauthoryear##1##2{##1}\@nbcite}

\maketitle

\begin{abstract}
Open-domain human-computer conversation has attracted much attention in the field of NLP. Contrary to rule- or template-based domain-specific dialog systems, open-domain conversation usually requires data-driven approaches, which can be roughly divided into two categories: retrieval-based and generation-based systems. Retrieval systems search a user-issued utterance (called a \textit{query}) in a large database, and return a reply that best matches the query. Generative approaches, typically based on recurrent neural networks (RNNs), can synthesize new replies, but they suffer from the problem of generating short, meaningless utterances. In this paper, we propose a novel ensemble of retrieval-based and generation-based dialog systems in the open domain. In our approach, the retrieved candidate, in addition to the original query, is fed to an RNN-based reply generator, so that the neural model is aware of more information. The generated reply is then fed back as a new candidate for post-reranking. Experimental results show that such ensemble outperforms each single part of it by a large margin.

\end{abstract}

\section{Introduction}

\sloppy
Automatic dialog/conversation systems have served humans for a long time in various fields, ranging from train routing~\cite{train} to museum guiding
~\cite{museum}. In the above scenarios, the dialogs are domain-specific, and a typical approach to such in-domain systems is by human engineering, for example, using manually constructed ontologies~\cite{youngsigdial}, natural language templates~\cite{template}, and even predefined dialog states~\cite{statetracking}.

Recently, researchers have paid increasing attention to open-domain, chatbot-style human-computer conversation, because of its important commercial applications, and because it tackles the real challenges of natural language understanding and generation~\cite{retrieval1,acl,aaai}.
For open-domain dialogs, rules and temples would probably fail as we can hardly handle the great diversity of dialog topics and natural language sentences.
With the increasing number of human-human conversation utterances available on the Internet, previous studies have developed data-oriented approaches in the open domain, which can be roughly categorized into two groups: retrieval systems and generative systems.

When a user issues an utterance (called a \textit{query}), retrieval systems search for a most similar query in a massive database (which consists of large numbers of query-reply pairs), and respond to the user with the corresponding reply \cite{retrieval1,retrieval2}. Through information retrieval, however, we cannot obtain new utterances, that is, all replies have to appear in the database. Also, the ranking of candidate replies is usually judged by surface forms (e.g., word overlaps, \textit{tf$\cdot$idf} features) and hardly addresses the real semantics of natural languages.

Generative dialog systems, on the other hand, can synthesize a new sentence as the reply by language models \cite{BoWdialog,acl,aaai}. Typically, a recurrent neural network (RNN) captures the query's semantics with one or a few distributed, real-valued vectors (also known as \textit{embeddings}); another RNN decodes the query embeddings to a reply.
Deep neural networks allow complicated interaction by multiple non-linear transformations;  RNNs are further suitable for modeling time-series data (e.g., a sequence of words) especially when enhanced with long short term memory (LSTM) or gated recurrent units (GRUs).
Despite these, RNN also has its own weakness when applied to dialog systems: the generated sentence tends to be short, universal, and meaningless, for example, ``I don't know''~\cite{naacl} or ``something''~\cite{aaai}. This is probably because chatbot-like dialogs are highly diversified and a query may not convey sufficient information for the reply.
Even though such universal utterances may be suited in certain dialog context, they make users feel boring and lose interest, and thus are not desirable in real applications.

In this paper, we are curious if we can combine the above two streams of approaches for open-domain conversation.
To this end, we propose an ensemble of retrieval and generative dialog systems. Given a user-issued query, we first obtain a candidate reply by information retrieval from a large database. The query, along with the candidate reply, is then fed to an utterance generator based on the ``bi-sequence to sequence'' (\multiseqseq) model~\cite{multiseq2seq}. Such sequence generator takes into consideration the information contained in not only the query but also the retrieved reply; hence, it alleviates the low-substance problem and can synthesize replies that are more meaningful.
After that we use the scorer in the retrieval system again
for post-reranking. This step can filter out less relevant retrieved replies or meaningless generated ones.
The higher ranked candidate (either retrieved or generated) is returned to the user as the reply.

From the above process, we see that the retrieval and generative systems are integrated by two mechanisms: (1) The retrieved candidate is fed to the sequence generator to mitigate the ``low-substance'' problem; (2) The post-reranker can make better use of both the retrieved candidate and the generated utterance. In this sense, we call our overall approach an \textit{ensemble} in this paper. To the best of our knowledge, we are the first to combine retrieval and generative models for open-domain conversation.

Experimental results show that our ensemble model consistently outperforms each single component in terms of several subjective and objective metrics, and that both retrieval and generative methods contribute an important portion to the overall approach.
This also verifies the rationale for building model ensembles for dialog systems.

\section{The Proposed Model Ensemble}
\sloppy
\subsection{Overview}

\begin{figure*}[!t]
\centering
\includegraphics[width=.9\textwidth]{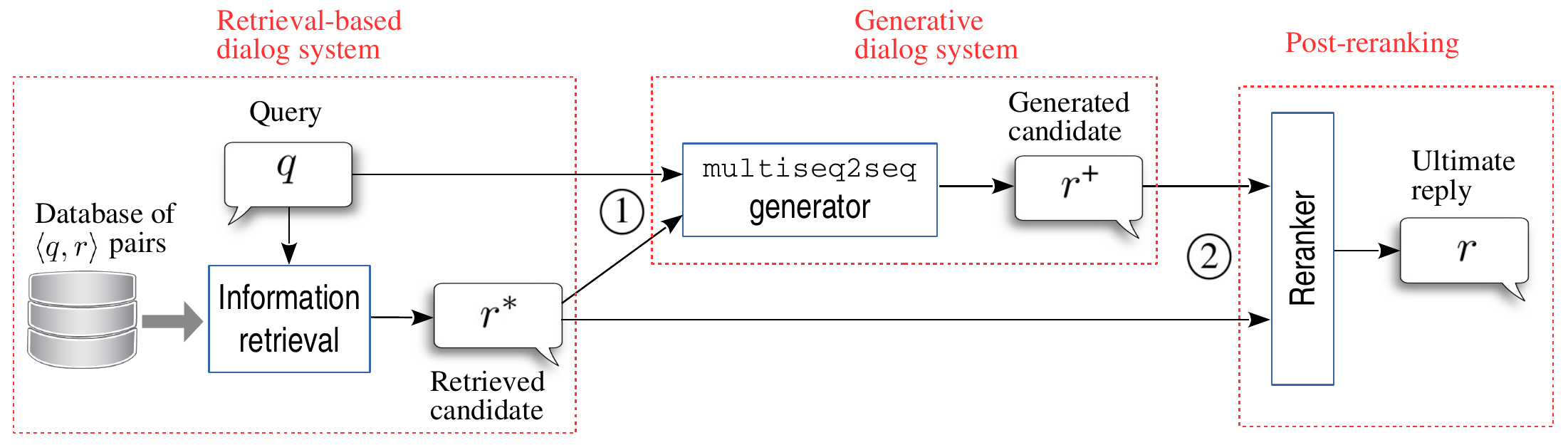}
\caption{The overall architecture of our model ensemble. We combine retrieval and generative dialog systems by \protect\circone\ enhancing the generator with the retrieved candidate and by \protect\circtwo\ post-reranking of both retrieved and generated candidates.}
\label{fig:overview}
\end{figure*}

Figure~\ref{fig:overview} depicts the overall framework of our proposed ensemble of retrieval and generative dialog systems. It mainly consists of the following components.

\begin{itemize}
\item When a user sends a query utterance $q$, our approach utilizes a state-of-the-practice information retrieval system to search for a query-reply pair $\langle q^*, r^*\rangle$ that best matches the user-issued query $q$. The corresponding $r^*$ is retrieved as a candidate reply.

\item Then a \multiseqseq\ model takes the original query $q$ and the retrieved candidate reply $r^*$ as input, each sequence being transformed to a fixed-size vector. These two vectors are concatenated and linearly transformed as the initial state of the decoder, which generates a new utterance $\rplus$ as another candidate reply.
\item Finally, we use a reranker (which is a part of the retrieval system) to select either $\rstar$ or $\rplus$ as the ultimate response to the original query $q$.
\end{itemize}
In the rest of this section, we describe each component in detail.

\subsection{Retrieval-Based Dialog System}

Information retrieval is among prevailing techniques for open-domain, chatbot-style human-computer conversation \cite{retrieval1,retrieval2}.

We utilize a state-of-the-practice retrieval system with extensive manual engineering and on a basis of tens of millions of existing human-human utterance pairs. Basically, it works in a two-step retrieval-and-ranking strategy, similar to the \texttt{Lucene}\footnote{http://lucene.apache.org} and \texttt{Solr}\footnote{http://lucene.apache.org/solr} systems.

First, a user-issued utterance is treated as bag-of-words features with stop-words being removed. After querying it in a knowledge base, we obtain a list containing up to 1000 query-reply pairs $\langle q^*, r^*\rangle$, whose queries share most words as the input query $q$.
This step retrieves coarse-grained candidates efficiently, which is accomplished by an inversed index.

Then, we measure the relatedness between the query $q$ and each $\langle q^*, r^*\rangle$ pair in a fine-grained fashion.
In our system, both $q$-$q^*$ and $q$-$r^*$ relevance scores are considered. A classifier judges whether $q$ matches $q^*$ and $r^*$; its confidence degree is used as the scorer. We have tens of features, and several important ones include word overlap ratio, the cosine measure of a pretrained topic model coefficients, and the cosine measures of word embedding vectors. 
(Details are beyond the scope of this paper; any well-designed retrieval system might fit into our framework.)

In this way, we obtain a query-reply pair $\langle q^*, r^*\rangle$ that best matches the original query $q$; the corresponding utterance $r^*$ is considered as a candidate reply retrieved from the database.

\subsection{The \multiseqseq\ Utterance Generator}

Using neural networks to build end-to-end trainable dialog systems has become a new research trend in the past year. A generative dialog system can synthesize new utterances, which is complementary  to retrieval-based methods.

Typically, an encoder-decoder architecture is applied to encode a query as vectors and to decode the vectors to a reply utterance. With recurrent neural networks (RNNs) as the encoder and decoder, such architecture is also known as a \seqseq\ model, which has wide applications in neural machine translation~\cite{seq2seq}, abstractive summarization~\cite{summarization},~etc. That being said, previous studies indicate \seqseq\ has its own shortcoming for dialog systems.\cite{seq2BF} suggests that, in open-domain conversation systems, the query does not carry sufficient information for the reply; that the \seqseq\ model thus tends to generate short and meaningless sentences with little substance.

\begin{figure}
\includegraphics[width=\linewidth]{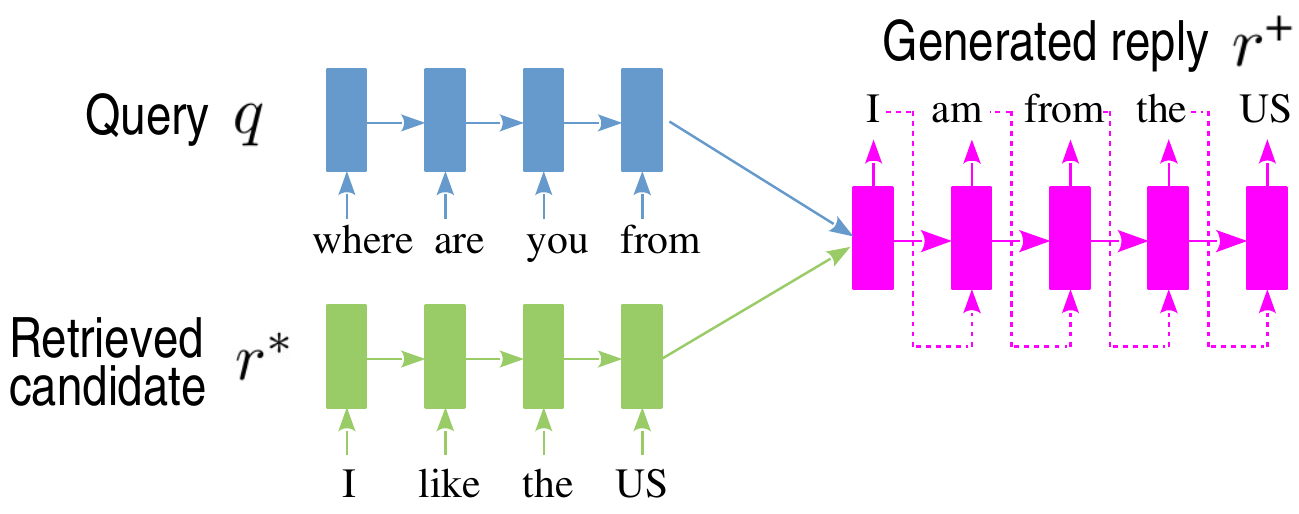}
\caption{The \multiseqseq\ model, which takes as input a query $q$ and a retrieved candidate reply $r^*$; it outputs a new reply $r^+$.}\label{fig:multiseqseq}
\end{figure}

To address this problem, we adopt a \multiseqseq\ model, which is proposed in \cite{multiseq2seq} for multi-source machine translation.
The \multiseqseq\ model takes into consideration the retrieved reply as a reference in addition to query information (Figure~\ref{fig:multiseqseq}). Hence, the generated reply can be not only fluent and logical with respect to the query, but also meaningful as it is enhanced by a retrieved candidate.

Specifically, we use an RNN with gated recurrent units (GRUs) for sequence modeling. Let $\bm x_t$ be the word embeddings of the time step $t$ and $\bm h_{t-1}$ be the previous hidden state of RNN. We have
\begin{align}\label{grubegin}
\bm r_t &= \sigma(W_r\bm x_t+ U_r\bm h_{t-1} + \bm b_r)\\
\bm z_t &= \sigma(W_z\bm x_t+ U_r\bm h_{t-1} + \bm b_z)\\
\tilde{\bm h}_t &= \tanh\big(W_x\bm x_t+ U_x (\bm r_t \circ \bm h_{t-1})\big)\\
\bm h_t &= (1-\bm z_t)\circ \bm h_{t-1} + \bm z_t \circ \tilde{\bm h}_t\label{gruend}
\end{align}
where $\bm r_t$ and $\bm z_t$ are known as gates, $W$'s and $\bm b$'s are parameters, and ``$\circ$'' refers to element-wise product.

After two RNNs go through $q$ and $r^*$, respectively, we obtain two vectors capturing their meanings. We denote them as bold letters $\bm q$ and $\bm r^*$, which are concatenated as $[\bm q; \bm r^*]$ and linearly transformed before being fed to the decoder as the initial state.

During reply generation, we also use GRU-RNN, given by Equations~\ref{grubegin}--\ref{gruend}. But at each time step, a softmax layer outputs the probability that a word would occur in the next step, i.e.,
\begin{equation}
p(w_i|\bm h_t) = \frac{\exp\left\{ W_i^\top \bm h_t\right\}+\bm b}
{\sum_j\exp\left\{ W_j^\top \bm h_t+\bm b\right\}}
\end{equation}
where $W_i$ is the $i$-th row of the output weight matrix (corresponding to $w_i$) and $\bm b$ is a bias term.

Notice that we assign different sets of parameters---indicated by three colors in Figure~\ref{fig:multiseqseq}---for the two encoders ($q$ and $r^*$) and the decoder ($r^+$). This treatment is because the RNNs' semantics differ significantly from one another (even between the two encoders).

\subsection{Post-Reranking}
Now that we have a retrieved candidate reply $r^*$ as well as a generated one $r^+$, we select one as the final reply by the $q$-$r$ scorer in the retrieval-based dialog system (described in previous sections and not repeated here).

Using manually engineered features, this step can eliminate either meaningless short replies that are
unfortunately generated by \multiseqseq\ or less relevant replies given by the retrieval system. We call this a \textit{post-reranker} in our model ensemble.

\subsection{Training}

We train each component separately because the retrieval part is not end-to-end learnable.

In the retrieval system, we use the classifier's confidence as the relevance score. The training set consists of 10k samples, which are either in the original human-human utterance pairs or generated by negative sampling. We made efforts to collect binary labels
from a crowd-sourcing platform, indicating whether a query is relevant to another query and whether it is relevant to a particular reply. We find using crowd-sourced labels results in better performance than original negative sampling.

For \multiseqseq, we use human-human utterance pairs $\langle q, r\rangle$ as data samples. A retrieved candidate $r^*$ is also provided as the input when we train the neural network.
Standard cross-entropy loss of all words in the reply is applied as the training objective. For a particular training sample whose reply is of length $T$, the cost is
\begin{equation}
J = -\sum_{i=1}^T\sum_{j=1}^{V}{t_j^{(i)}\log{y_j^{(i)}}}
\end{equation}
where $\bm{t}^{(i)}$ is the one-hot vector of the next target word, serving as the groundtruth, $\bm y$ is the output probability by softmax, and $V$ is the vocabulary size.
We adopt mini-batched AdaDelta~\cite{adadelta} for optimization.

\section{Evaluation}
\sloppy
In this section, we evaluate our model ensemble on Chinese (language) human-computer conversation. We first describe the datasets and settings. Then we compare our approach with strong baselines.

\subsection{Experimental Setup}
Typically, a very large database of query-reply pairs is a premise for a successful retrieval-based conversation system, because the reply must appear in the database. For RNN-based sequence generators, however, it is time-consuming to train with such a large dataset; RNN's performance may also saturate when we have several million samples. 

\begin{table}[!t]
\centering
\renewcommand\arraystretch{1.1}
\begin{tabular}{|l|r|}
\hline
\multicolumn{1}{|c|}{\textbf{Dataset Split}}           & \quad\textbf{\# of samples}\\
\hline
\hline
 Retrieval (Database)\quad\quad  &\quad 7,053,820  \\
 Matching scorer (Train)     & 50,000\\
 Generator (Train)\quad & 1,500,000   \\
\hline
\multicolumn{1}{|l|}{ Validation}       & 100,000  \\
\multicolumn{1}{|l|}{ Testing}           & 6,741 \\
\hline
\end{tabular}
\caption{Statistics of our datasets.}
\end{table}
To construct a database for information retrieval, we collected human-human utterances from massive online forums, microblogs, and question-answering communities, such as Sina Weibo,\footnote{http://weibo.com} Baidu Zhidao,\footnote{http://zhidao.baidu.com} and Baidu Tieba.\footnote{http://tieba.baidu.com}
We filtered out short and meaningless replies like ``\dots'' and ``Errr.''
In total, the database contains 7 million query-reply pairs for retrieval.

For the generation part, we constructed another dataset from various resources in public websites comprising 1,606,741 query-reply pairs. For each query $q$, we searched for a candidate reply $r^*$ by the retrieval component and obtained a tuple $\langle q, r^*, r\rangle$. As a friendly reminder, $q$ and $r^*$ are the input of \multiseqseq, whose output should approximate~$r$.
We randomly selected 100k triples for validation and another 6,741 for testing. The train-val-test split remains the same for all competing models.

The \multiseqseq\ then degrades to an utterance autoencoder~\cite{autoencoder}. Also, the validation and test sets are disjoint with the training set and the database for retrieval, which complies with the convention of machine learning.

To train our neural models, we followed \cite{acl} for hyperparameter settings. All embeddings were set to 620-dimensional and  hidden states 1000d.  We applied AdaDelta with a mini-batch size of 80 and other default hyperparameters for optimization. Chinese word segmentation was performed on all utterances.
We kept a same set of 100k words (Chinese terms) for two encoders, but 30k for the decoder due to efficiency concerns.
The three neural networks do not share parameters (neither connection weights nor embeddings).

We did not tune the above hyperparameters, which were set empirically. The validation set was used for early stop based on the perplexity measure.

\begin{table*}[!t]
\centering
\begin{tabular}{|l||c||r|r|r|r|}
\hline
\textbf{Method}              & \textbf{Human Score} & \textbf{BLEU-1}  & \textbf{BLEU-2}
& \textbf{BLEU-3} & \textbf{BLEU-4} \\\hline
\hline
Retrieval       & 0.996    & 5.707     & 3.092   & 2.406 & \textbf{2.094}  \\
\seqseq\        & 0.907    & 3.676     &  1.228    & 0.564 &0.286 \\
\hline
\multiseqseq\   & 0.966    & 7.762     & 3.293   & 2.056 &1.487 \\
Rerank(Retrieval,\seqseq) & 1.030 & 4.500 & 2.041 & 1.364  & 1.060\\
Rerank(Retrieval,\multiseqseq) & \textbf{1.131} & \textbf{7.260} & \textbf{3.503} & \textbf{2.480} & 2.000\\
\hline
\end{tabular}
\vspace{-.1cm}
\caption{Results of our ensemble and competing methods in terms of average human scores and BLEUs.  Inter-annotator agreement for human annotation: Fleiss' $\kappa=0.2824$~\cite{kappa}, std $=0.4031$, indicating moderate agreement.}
\label{tab:result}
\vspace{-.2cm}
\end{table*}

\subsection{Competing Methods}
We compare our model ensemble with each individual component and provide a thorough ablation test. Listed below are the competing methods in our experiments.
\begin{itemize}
\item \textit{Retrieval}. A state-of-the-practice dialog system, which is a component of our model ensemble; it is also a strong baseline because of extensive human engineering.
\item \textit{\seqseq}. An encoder-encoder framework~\cite{seq2seq}, first introduced in~\cite{acl} for dialog systems.
\item \textit{\multiseqseq}. Another component in our approach, adapted from~\cite{multiseq2seq}, which is essentially a \textit{\seqseq} model extended with a retrieved reply.
\item \textit{Rerank(Retrieval,\seqseq)}. Post-reranking between a retrieved candidate and one generated by \seqseq.
\item\textit{Rerank(Retrieval,\multiseqseq)}. This is the full proposed model ensemble.
\end{itemize}

All baselines were trained and tuned in a same way as our full model, when applicable, so that the comparison is fair.

\subsection{Overall Performance}
We evaluated our approach in terms of both subjective and objective evaluation.
\begin{itemize}
\item Human evaluation, albeit time- and labor-consuming, conforms to the ultimate goal of open-domain conversation systems. We asked three educated volunteers to annotate the results using a common protocol known as \textit{pointwise annotation} \cite{acl,ijcai,seq2BF}. In other words, annotators were asked to label either ``0'' (bad), ``1'' (borderline), or ``2'' (good) to a query-reply pair. The subjective evaluation was performed in a strict random and blind fashion to rule out human bias.
\item We adopted BLEU-1, BLEU-2, BLEU-3 and BLEU-4 as automatic evaluation. While \cite{howNOTto} further aggressively argues that no existing automatic metric is appropriate for open-domain dialogs, they show a slight positive correlation between BLEU-2 and human evaluation in non-technical Twitter domain, which is similar to our scenario. We nonetheless include BLEU scores as expedient objective evaluation, serving as supporting evidence. BLEUs are also used in~\cite{naacl} for model comparison and in~\cite{seq2BF} for model selection.
\end{itemize}
Notice that, automatic metrics were computed on the entire test set, whereas subjective evaluation was based on 79 randomly chosen test samples due to the limitation of human resources available.

We present our main results in Table~\ref{tab:result}. As shown, the retrieval system, which our model ensemble is based on, achieves better performance than RNN-based sequence generation.
The result is not consistent with~\cite{acl}, where their RNNs are slightly better than retrieval-based methods. After closely examining their paper, we find that their database is multiple times smaller than ours, which may, along with different features and retrieval methods, explain the phenomenon. This also verifies that the retrieval-based dialog system in our experiment is a strong baseline to compare with.

Combining the retrieval system and the RNN generator by bi-sequence input and post-reranking, we achieve the highest performance in terms of both human evaluation and BLEU scores. Concretely, our model ensemble outperforms the state-of-the-practice retrieval system by $ +13.6\%$ averaged human scores, which we believe is a large margin.

\subsection{Analysis and Discussion}
\begin{table}[!t]
\renewcommand\arraystretch{1.1}
\centering
\begin{tabular}{|l|c|c|}
\hline
\textbf{Method}\quad\quad\quad\quad\quad\quad        & \textbf{Entropy}  & \textbf{Length}\\
\hline
\hline
\seqseq      & 7.420 & 7.362\\
\multiseqseq & \textbf{8.302} & \textbf{8.185}\\
\hline
Groundtruth & 8.625  & 12.62
 \\
\hline
\end{tabular}
\caption{Entropy and length of generated replies. We also include groundtruth for reference. A larger entropy value indicates that the replies are less common, and probably, more meaningful.}
\label{tab:entropy}
\vspace{-.2cm}
\end{table}
Having verified that our model ensemble achieves better performance than all baselines, we are further curious how each gadget contributes to our final system. Specially, we focus on the following research questions.

\medskip
\noindent{\textbf{RQ1}}: What is the performance of \multiseqseq\ (the \circone\ step in Figure~\ref{fig:overview}) in comparison with traditional \seqseq?

\smallskip
From the BLEU scores in Table~\ref{tab:result}, we see \multiseqseq\ significantly outperforms conventional \seqseq, showing that, if enriched with a retrieved human utterance as a candidate, the encoder-decoder framework can generate much more human-like utterances.

We then introduce in Table~\ref{tab:entropy} another measure, the entropy of a sentence, defined as $$-\frac1{|R|}\sum_{w\in R}\log p(w)$$
where $R$ refers to all replies. Entropy is used in \cite{variationalDialog} and \cite{seq2BF} to measure the serendipity of generated utterances.\footnote{Notice that, the entropy of retrieved replies is not a fair metric to compare in Table~\ref{tab:entropy}, because the retrieval system has filtered out short, meaningless utterances in advance by surface statistics (e.g., length). We nevertheless report the result here out of curiosity: its entropy is 9.507, which is even higher than groundtruth.}
The results in Table~\ref{tab:entropy} confirm that  \multiseqseq\ indeed integrates information from the retrieved candidate, so that it alleviates the ``low-substance'' problem of RNNs and  can generate  utterances more meaningful than traditional \seqseq. And the statistic result also displays that \multiseqseq\ generates longer sentences than \seqseq\ approach.

\begin{figure}[!t]
\centering
\includegraphics[width=.9\linewidth]{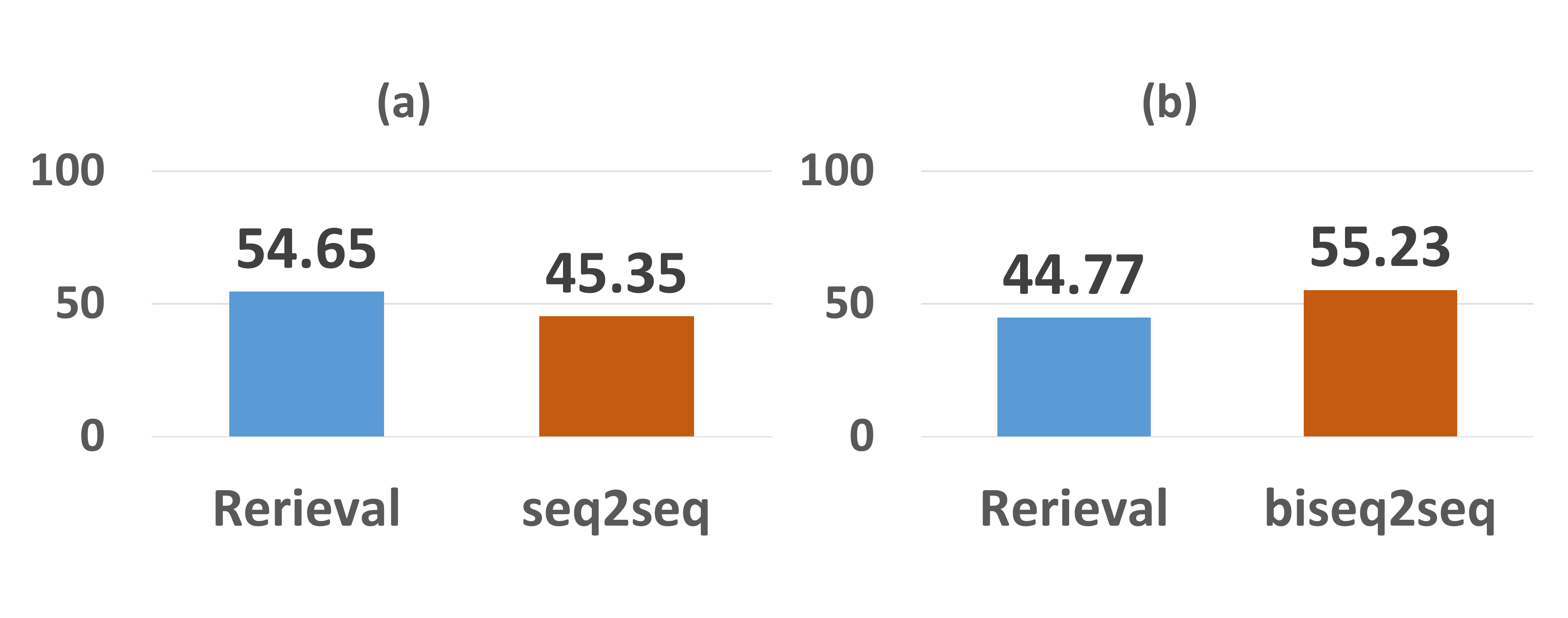}
\vspace{-.3cm}
\caption{The percentage by which our post-reranker chooses a retrieved reply or a generated reply. (a) Ensemble of \textit{Retrieval} and \textit{\seqseq}; (b) Ensemble of \textit{Retrieval} and \textit{\multiseqseq}.}
\vspace{-.2cm}
\label{fig:rerank}
\end{figure}

\begin{table*}[!t]
\centering
\begin{tabular}{|l|l|l|}
\hline
\Tsmall
 & \textbf{Utterance  (Translated)} & \textbf{Reranked} \\
 \hline\hline\Tstrut
\multirow{2}{*}{Query} & 手机拍照效果不错 ~ & \cellcolor{mygray} \\
&(This mobile phone's photo effect is pretty good. )& \cellcolor{mygray}\\
\cline{1-2}\Tstrut
{Retrieved} &  所以好心动~(I really have a crush on it.) &\\
\cline{1-2}\Tstrut
\multiseqseq &心动不如行动~(Rushing for it rather than having a crush on it.)&  $\leftarrow$ \\
\cline{1-2}\Tstrut
\seqseq & 哈哈~(Haha) &\cellcolor{mygray}\\
\hline\hline\Tstrut
\multirow{2}{*}{Query}        &  有给孕妇推荐的保健品啥的么？~ & \cellcolor{mygray} \\
 & (Could you please recommend some health products for pregnant woman? ) &  \cellcolor{mygray}   \\
\cline{1-2}\Tstrut
{Retrieved}    & 打算上孕妇装？~(Do you want metarnity clothes? ) & \\
\cline{1-2} \Tstrut
\multiseqseq & 打算买护肤品么？~(Do you want some skin care products?)     &    $\leftarrow$    \\
\cline{1-2} \Tstrut
\seqseq      & 不好意思~(I'm sorry. )     &     \cellcolor{mygray}    \\
\hline
\end{tabular}
\vspace{-.1cm}
\caption{Examples of retrieved replies and generated ones. An arrow ``$\leftarrow$'' indicates the one selected during post-reranking. Also included are replies generated by conventional \seqseq. Notice that it is not part of our model and thus not considered for reranking.}\label{tab:case}
\vspace{-.2cm}
\end{table*}
\medskip
\noindent{\textbf{RQ2}}: How do the retrieval- and generation-based systems contribute to  post-reranking (the \circtwo\ step in Figure~\ref{fig:overview})?

\smallskip
We plot in Figure~\ref{fig:rerank} the percentage by which the post-raranker chooses a retrieved candidate or a generated one. In the retrieval-and-\seqseq\ ensemble (Figure~\ref{fig:rerank}a), 54.65\% retrieved results and 45.35\% generated ones are selected. In retrieval-and-\multiseqseq\ ensemble, the percentage becomes 44.77\% vs.~55.23\%. The trend further indicates that \multiseqseq\ is better than \seqseq\ (at least) from the reranker's point of view. More importantly, as the percentages are close to 50\%, both the retrieval system and the generation system contribute a significant portion to our final ensemble.

\bigskip
\noindent{\textbf{RQ3}}: Do we obtain further gain by combining the two gadgets \circone\ and \circtwo\ in Figure~\ref{fig:overview}?

\smallskip
We would also like to verify if the combination of \multiseqseq\ and post-reranking mechanisms will yield further gain in our ensemble. To test this, we compare the full model \textit{Rerank(Retrieval,\multiseqseq)} with an ensemble that uses traditional \seqseq, i.e., \textit{Rerank(Retrieval,\seqseq)}.
As indicated in Table~\ref{tab:result}, even with the post-reranking mechanism, the ensemble with underlying \multiseqseq\ still outperforms the one with \seqseq.
Likewise, \textit{Rerank(Retrieval,\multiseqseq)} outperforms both \textit{Retrieval} and \textit{\multiseqseq}. These results are consistent in terms of all metrics except a BLEU-4 score.

Through the above ablation tests, we conclude that both gadgets (\multiseqseq\ and post-reranking) play a role in our ensemble when we combine the retrieval and the generative systems.

\subsection{Case Study}
Table~\ref{tab:case} presents two examples of our ensemble and its ``base'' models. We see that \multiseqseq\ is indeed influenced by the retrieved candidates. As opposed to traditional \seqseq, several content words in the retrieved replies (e.g., \textit{crush}) also appear in \multiseqseq's output, making the utterances more meaningful. The post-reranker also chooses a more appropriate candidate as the reply.

\section{Related Work}
\sloppy
In early years, researchers mainly focus on domain-specific dialog systems, e.g., train routing~\cite{train}, movie information~\cite{movie}, and human tutoring~\cite{tutor}.
Typically, a pre-constructed ontology defines a finite set of slots and values, for example, cuisine, location, and price range in a food service dialog system; during human-computer interaction, a state tracker fills plausible values to each slot from user input, and recommend the restaurant that best meets the user's requirement~\cite{webstyle,ACL15statetracking,pseudoN2N}.

In the open domain, however, such slot-filling approaches would probably fail because of the diversity of topics and natural language utterances. \cite{retrieval1} applies information retrieval techniques to search for related queries and replies. \cite{retrieval2} and \cite{sigir} use both shallow hand-crafted features and deep neural networks for matching.
\cite{ijcai} proposes a random walk-style algorithm to rank candidate replies. In addition, their model can introduce additional content (related entities in the dialog context) by searching a knowledge base when a stalemate occurs during human-computer conversation.

Generative dialog systems have recently attracted increasing attention in the NLP community.
\cite{smt} formulates query-reply transformation as a phrase-based machine translation.
Since the last year, the renewed prosperity of neural networks witnesses an emerging trend in using RNN for dialog systems~\cite{nn0,BoWdialog,acl,aaai}. However, a known issue with RNN is that it prefers to generate short, meaningless utterances. \cite{naacl} proposes a mutual information objective in contrast to the conventional maximum likelihood criterion. \cite{seq2BF} and \cite{topic} introduce additional content (either the most mutually informative word or topic information) to the reply generator. \cite{variationalDialog} applies a variational encoder to capture query information as a distribution, from which  a random vector is sampled for reply generation.

To the best of our knowledge, we are the first to combine retrieval-based and generation-based dialog systems. The use of \multiseqseq\ and post-reranking is also a new insight of this paper.

\section{Conclusion and Future Work}
\sloppy

In this paper, we propose a novel ensemble of retrieval-based and generation-based open-domain dialog systems. The retrieval part searches a best-match candidate reply, which is, along with the original query, fed to an RNN-based \multiseqseq\ reply generator. The generated utterance is fed back as a new candidate  to the retrieval system for post-reranking.
Experimental results show that our ensemble outperforms its underlying retrieval system and generation system by a large margin. In addition, the ablation test demonstrates both the \multiseqseq\ and post-reranking mechanisms play an important role in the ensemble.

Our research also points out several promising directions for future work, for example, developing new mechanisms of combining retrieval and generative dialog systems, as well as incorporating other data-driven approaches to human-computer conversation.

\bibliographystyle{plain}
\bibliography{main}

\begin{thebibliography}{10}

\bibitem{movie}
Jennifer Chu-Carroll.
\newblock {MIMIC: A}n adaptive mixed initiative spoken dialogue system for
  information queries.
\newblock In {\em Proc. Conf. Applied Natural Language Processing}, pages
  97--104, 2000.

\bibitem{train}
G.~Ferguson, J.~Allen, B.~Miller, et~al.
\newblock Trains-95: Towards a mixed-initiative planning assistant.
\newblock In {\em AIPS}, pages 70--77, 1996.

\bibitem{kappa}
Joseph~L Fleiss.
\newblock Measuring nominal scale agreement among many raters.
\newblock {\em Psychological Bulletin}, 76(5):378, 1971.

\bibitem{museum}
Nadine Glas, Ken Prepin, and Catherine Pelachaud.
\newblock Engagement driven topic selection for an information-giving agent.
\newblock In {\em Proc. Workshop on the Semantics and Pragmatics of Dialogue},
  2015.

\bibitem{tutor}
A.~Graesser, P.~Chipman, B.~Haynes, and A.~Olney.
\newblock {AutoTutor: A}n intelligent tutoring system with mixed-initiative
  dialogue.
\newblock {\em IEEE Trans. Education}, 48(4):612--618, 2005.

\bibitem{retrieval1}
Charles~Lee Isbell, Michael Kearns, Dave Kormann, Satinder Singh, and Peter
  Stone.
\newblock {Cobot in LambdaMOO: A} social statistics agent.
\newblock In {\em AAAI}, pages 36--41, 2000.

\bibitem{retrieval2}
Zongcheng Ji, Zhengdong Lu, and Hang Li.
\newblock An information retrieval approach to short text conversation.
\newblock {\em arXiv preprint arXiv:1408.6988}, 2014.

\bibitem{naacl}
Jiwei Li, Michel Galley, Chris Brockett, Jianfeng Gao, and Bill Dolan.
\newblock A diversity-promoting objective function for neural conversation
  models.
\newblock In {\em NAACL-HLT}, pages 110--119, 2016.

\bibitem{autoencoder}
Jiwei Li, Thang Luong, and Dan Jurafsky.
\newblock A hierarchical neural autoencoder for paragraphs and documents.
\newblock In {\em ACL-IJCNLP}, pages 1106--1115, 2015.

\bibitem{ijcai}
Xiang Li, Lili Mou, Rui Yan, and Ming Zhang.
\newblock {StalemateBreaker: A} proactive content-introducing approach to
  automatic human-computer conversation.
\newblock In {\em IJCAI}, pages 2845--2851, 2016.

\bibitem{howNOTto}
Chia-Wei Liu, Ryan Lowe, Iulian~V Serban, Michael Noseworthy, Laurent Charlin,
  and Joelle Pineau.
\newblock How {NOT} to evaluate your dialogue system: An empirical study of
  unsupervised evaluation metrics for dialogue response generation.
\newblock In {\em EMNLP (to appear)}, 2016.

\bibitem{seq2BF}
Lili Mou, Yiping Song, Rui Yan, Ge~Li, Lu~Zhang, and Zhi Jin.
\newblock Sequence to backward and forward sequences: A content-introducing
  approach to generative short-text conversation.
\newblock {\em arXiv preprint arXiv:1607.00970}, 2016.

\bibitem{ACL15statetracking}
Nikola Mrk\v{s}i\'{c}, Diarmuid \'{O}~S\'{e}aghdha, Blaise Thomson, Milica
  Gasic, Pei-Hao Su, David Vandyke, Tsung-Hsien Wen, and Steve Young.
\newblock Multi-domain dialog state tracking using recurrent neural networks.
\newblock In {\em ACL-IJCNLP}, pages 794--799, 2015.

\bibitem{smt}
Alan Ritter, Colin Cherry, and William~B Dolan.
\newblock Data-driven response generation in social media.
\newblock In {\em EMNLP}, pages 583--593, 2011.

\bibitem{summarization}
Alexander~M. Rush, Sumit Chopra, and Jason Weston.
\newblock A neural attention model for abstractive sentence summarization.
\newblock In {\em EMNLP}, pages 379--389, 2015.

\bibitem{aaai}
Iulian~V Serban, Alessandro Sordoni, Yoshua Bengio, Aaron Courville, and Joelle
  Pineau.
\newblock Building end-to-end dialogue systems using generative hierarchical
  neural network models.
\newblock In {\em AAAI}, pages 3776--3783, 2016.

\bibitem{variationalDialog}
Iulian~Vlad Serban, Alessandro Sordoni, Ryan Lowe, Laurent Charlin, Joelle
  Pineau, Aaron Courville, and Yoshua Bengio.
\newblock A hierarchical latent variable encoder-decoder model for generating
  dialogues.
\newblock {\em arXiv preprint arXiv:1605.06069}, 2016.

\bibitem{acl}
Lifeng Shang, Zhengdong Lu, and Hang Li.
\newblock Neural responding machine for short-text conversation.
\newblock In {\em ACL-IJCNLP}, pages 1577--1586, 2015.

\bibitem{BoWdialog}
Alessandro Sordoni, Michel Galley, Michael Auli, Chris Brockett, Yangfeng Ji,
  Margaret Mitchell, Jian-Yun Nie, Jianfeng Gao, and Bill Dolan.
\newblock A neural network approach to context-sensitive generation of
  conversational responses.
\newblock In {\em NAACL-HLT}, pages 196--205, 2015.

\bibitem{template}
Pei-Hao Su, Milica Gasic, Nikola Mrksic, Lina Rojas-Barahona, Stefan Ultes,
  David Vandyke, Tsung-Hsien Wen, and Steve Young.
\newblock Continuously learning neural dialogue management.
\newblock {\em arXiv preprint arXiv:1606.02689}, 2016.

\bibitem{seq2seq}
Ilya Sutskever, Oriol Vinyals, and Quoc~V Le.
\newblock Sequence to sequence learning with neural networks.
\newblock In {\em NIPS}, pages 3104--3112, 2014.

\bibitem{nn0}
Oriol Vinyals and Quoc Le.
\newblock A neural conversational model.
\newblock {\em arXiv preprint arXiv:1506.05869}, 2015.

\bibitem{youngsigdial}
Tsung-Hsien Wen, Milica Gasic, Dongho Kim, Nikola Mrksic, Pei-Hao Su, David
  Vandyke, and Steve Young.
\newblock Stochastic language generation in dialogue using recurrent neural
  networks with convolutional sentence reranking.
\newblock In {\em SIGDIAL}, pages 275--284, 2015.

\bibitem{pseudoN2N}
Tsung-Hsien Wen, Milica Gasic, Nikola Mrksic, Lina~M Rojas-Barahona, Pei-Hao
  Su, Stefan Ultes, David Vandyke, and Steve Young.
\newblock A network-based end-to-end trainable task-oriented dialogue system.
\newblock {\em arXiv preprint arXiv:1604.04562}, 2016.

\bibitem{statetracking}
Jason Williams, Antoine Raux, Deepak Ramachandran, and Alan Black.
\newblock The dialog state tracking challenge.
\newblock In {\em SIGDIAL}, pages 404--413, 2013.

\bibitem{webstyle}
Jason~D Williams.
\newblock Web-style ranking and {SLU} combination for dialog state tracking.
\newblock In {\em SIGDIAL}, pages 282--291, 2014.

\bibitem{topic}
Chen Xing, Wei Wu, Yu~Wu, Jie Liu, Yalou Huang, Ming Zhou, and Wei-Ying Ma.
\newblock Topic augmented neural response generation with a joint attention
  mechanism.
\newblock {\em arXiv preprint arXiv:1606.08340}, 2016.

\bibitem{sigir}
Rui Yan, Yiping Song, and Hua Wu.
\newblock Learning to respond with deep neural networks for retrieval-based
  human-computer conversation system.
\newblock In {\em SIGIR}, pages 55--64, 2016.

\bibitem{adadelta}
Matthew~D Zeiler.
\newblock {AdaDelta: A}n adaptive learning rate method.
\newblock {\em arXiv preprint arXiv:1212.5701}, 2012.

\bibitem{multiseq2seq}
Barret Zoph and Kevin Knight.
\newblock Multi-source neural translation.
\newblock In {\em NAACL-ACL}, pages 30--34, 2016.

\end{thebibliography}

\end{CJK*}
\end{document}